\documentclass[times, twoside]{zHenriquesLab-StyleBioRxiv-Preprint}

\usepackage{blindtext}
\usepackage{graphicx}
\usepackage{dblfloatfix} 
\usepackage{lineno}
\usepackage{comment}
\usepackage[normalem]{ulem}
\usepackage{color, colortbl}
\usepackage{MnSymbol}

\leadauthor{Ballester, Wang} 

\begin{document}

\setlength{\parindent}{1cm}

\title{Single-shot ToF sensing with sub-mm precision using conventional CMOS sensors}

\shorttitle{Main Paper}


\author[1,*]{Manuel Ballester}
\author[2,*]{Heming Wang}
\author[2]{Jiren Li}
\author[1,2]{Oliver Cossairt}
\author[3,**]{Florian Willomitzer}

\affil[1]{Department of Computer Sciences, Northwestern University, Evanston, IL 60208}
\affil[2]{Department of Electrical and Computer Engineering, Northwestern University, Evanston, IL 60208}
\affil[3]{Wyant College of Optical Sciences, University of Arizona, Tucson, AZ 85721}
\affil[*]{Joint first authorship}
\affil[**]{Correspondence: fwillomitzer@arizona.edu}

\maketitle

\begin{abstract}

We present a novel single-shot interferometric  ToF camera targeted for precise 3D measurements of dynamic objects. The camera concept is based on Synthetic Wavelength Interferometry, a technique that allows retrieval of depth maps of objects with optically rough surfaces at submillimeter depth precision. In contrast to conventional ToF cameras, our device uses only off-the-shelf CCD/CMOS detectors and works at their native chip resolution (as of today, theoretically up to 20 Mp and beyond). Moreover, we can obtain a full 3D model of the object in single-shot, meaning that no temporal sequence of exposures or temporal illumination modulation (such as amplitude or frequency modulation) is necessary, which makes our camera robust against object motion.

In this paper, we introduce the novel camera concept and show first measurements that demonstrate the capabilities of our system. We present 3D measurements of small (cm-sized) objects with > 2 Mp point cloud resolution (the resolution of our used detector) and up to sub-mm depth precision. We also report a ``single-shot 3D video'' acquisition and a first single-shot ``Non-Line-of-Sight'' measurement. Our technique has great potential for high-precision applications with dynamic object movement, e.g., in AR/VR, industrial inspection, medical imaging, and imaging through scattering media like fog or human tissue. 

\end {abstract}

\vspace{-3mm}

\section{Introduction}
\label{sec:intro}

3D imaging techniques are progressively replacing conventional photographs, which cannot capture the complexity of real-world objects. A 3D model provides information about an object's surface shape, slope, or texture - essential properties in object analysis. Once we obtain a realistic 3D model of an object, we can reproduce it under different illuminations and rigid transformations using computer graphics techniques. These features make 3D imaging very useful for many applications, including medical diagnostic imaging, quality inspection of manufactured products, and autonomous driving.

There are numerous 3D imaging techniques in the literature, which can be mainly classified into three groups. The first group corresponds to triangulation-based techniques, which are one of the most commonly used techniques in industrial inspection or 3D metrology on macroscopic rough surfaces. Examples include active and passive stereo, line- or fringe-projection techniques, or focus-searching methods \cite{trian1, trian2, trian4, trian5, trian7}. However, the depth precision of related methods scales with the stand-off distance, and occluded parts of the object cannot be measured. The second group refers to all reflectance-based methods, which typically use specific lighting conditions and then capture the light reflected or scattered off the object's surface to estimate its surface slope. Some examples include photometric stereo and deflectometry \cite{reflection1, reflection2, reflection3, reflection5, reflection6, reflection7}. These systems often become highly complicated to calibrate, and the integration from the surface normal to surface shape is problematic in some cases. The third category corresponds to all Time-of-Flight-based  (ToF) approaches \cite{book, ToF16, heide}, that calculate the optical pathlength between camera and object or scene to obtain its shape.  ToF cameras of all types (pulsed, amplitude modulated, etc) and optical interferometers are prominent examples for this group. While conventional ToF cameras directly measure the travel time of light, interferometric cameras measure the optical pathlength difference via interference patterns produced by an object and a reference beam. ToF-based approaches are beneficial for many applications because they are robust to occlusions and the data precision does not depend on the standoff distance of the camera. 

As mentioned, conventional ``ToF cameras" typically exploit amplitude-modulated light sources either as pulsed illumination (LIDAR) \cite{lidar1, lidar2} or in the form of ``continuous waves with amplitude modulation" (CWAM) \cite{cwam1, cwam2, cwam3}. These cameras generally require specific sensor architectures with a high temporal resolution. The main restriction of these cameras is their (still) fairly limited pixel resolution and their low depth precision, which is roughly on the order of centimeters. This makes conventional ToF cameras useful for basic estimation or detection tasks of larger objects (for example, to know if there is an object in front of an autonomous driving vehicle, or to estimate the pose of a human). However, they cannot be used  for applications that require high data precision, such as specific tasks in medical imaging, industrial inspection, or optical metrology. Moreover, some ToF camera schemes require a sequence of exposures to calculate a 3D scene or exploit raster-scanning of a laser dot (i.e., are not ``single-shot"). That makes them susceptible to fast object motions.

On the other hand, the main drawback of standard
(single-wavelength) interferometric ToF cameras are that
they are limited to the measurement of objects with optically
smooth (specular) surfaces \cite{hariharan, interferometry,born}. When an optically rough surface is illuminated with coherent light, the backscattered  light field forms a ``speckle pattern". Since the phase of this speckle field is randomized, we cannot recover information about the optical pathlength difference and hence, the depth map of the object. In practice, standard interferometric techniques become helpful only in particular fields, e.g., to study the nanomechanical motion of some objects, or for high-precision surface testing of lenses or smooth technical parts. 

Nevertheless, multifrequency interferometric ToF cameras \cite{interferometry, fercher1, fercher2, dandliker1, dandliker2, Groot1, Groot2} have shown that the interferometric principles can also extend to macroscopic objects with rough surfaces. These methods use the information from two (or more) interferometric measurements at different optical wavelengths to disambiguate the  \textit{random-phase fluctuations} of the object beam. Nevertheless, recent implementations of these techniques  for Computer Vision applications \cite{ICCP, PAMI, SWI2, Ioannis, book} still require sophisticated sensors or are limited to static objects (that is, they are not motion-robust).

In this contribution, we present a novel multifrequency interferometric ToF camera concept and device that overcomes the described limitations of the state-of-the-art (SOTA) ToF cameras shown above. Our demonstrated camera prototype scans object surfaces in single-shot with up to submillimeter  depth precision. Here, ``single-shot" means that only one camera image is necessary to generate a full-field high-quality 3D model. This feature makes our ToF camera robust against object motion. Moreover, the camera only uses off-the-shelf CMOS or CCD detector technology, that is, it theoretically even works with standard smartphone cameras. No specialized detector architecture (like PMDs \cite{cwam1}, SPADs \cite{SPAD}, or lock-in detection \cite{lock-in}) is required. The lateral point cloud resolution of our camera is equal to the pixel resolution of the used CMOS/CCD chip - theoretically up to 20Mp and beyond, given the current state of the art in CMOS camera technology \cite{CMOS}. These capabilities give our camera significant advantages with respect to the current SOTA in ToF camera technology and allow for many high-precision 3D imaging applications that previously were not possible with ToF cameras. We summarize  the specific contributions of our work as follows:

\begin{itemize}
\setlength{\itemsep}{0pt}%
\item We devised and developed a novel ToF camera technology. This technology is based on a previously introduced method called ``Synthetic Wavelength Interferometry" (SWI) \cite{book, fercher1, fercher2, dandliker1, dandliker2, Groot1, Groot2}. Compared to previous demonstrations of SWI for Computer Vision applications \cite{ICCP, PAMI, SWI2, Ioannis}, our method allows to scan the object surface in single-shot, using only off-the-shelf CMOS/CCD camera technology.

\item  We demonstrate ToF-based 3D measurements of static objects with rough surfaces with up to $330 \mu m$ depth precision and 2 Megapixel lateral point cloud resolution (= the pixel resolution of the used CMOS camera). The measurements shown in Figs. \ref{fig:double} and \ref{fig:single} exploit multifrequency phase unwrapping procedures \cite{multifreq} to increase the depth sensitivity of the sensing principle, which means that these measurements are not single-shot. Later, Figs. \ref{fig:metronome} and \ref{fig:DL} show real single-shot measurements. Finally, we demonstrate the 3D video reconstruction of a moving object from sequential single-shot measurements. To the best of our knowledge, this is the first demonstration of an SWI-based video sequence for Computer Vision applications.

\item To combine the high-depth precision of multifrequency measurements with the motion robustness of single-shot measurements, we have developed a deep-learning-based algorithm for single-shot phase unwrapping. Compared to other SOTA phase unwrapping procedures \cite{unwrap_heide1, unwrap_heide2, unwrap_probabilistic, unwrap_maxlikelihood}, our algorithm is specifically tailored to our acquired data structures and integrates seamlessly into our data processing pipeline. Although still a work in progress, we demonstrate the first single-shot measurements unwrapped with our algorithm.

\item We use our novel single-shot technique to acquire a ``Non-Line-of-Sight" measurement of a point object hidden behind a scatterer. To the best of our knowledge, this experiment documents the first ever SWI-based single-shot measurement through scattering media.

\end{itemize}

\section{Principles and computational framework}
\label{sec:math}

\begin{figure*}[t!]
  \centering
   \includegraphics[width=\linewidth]{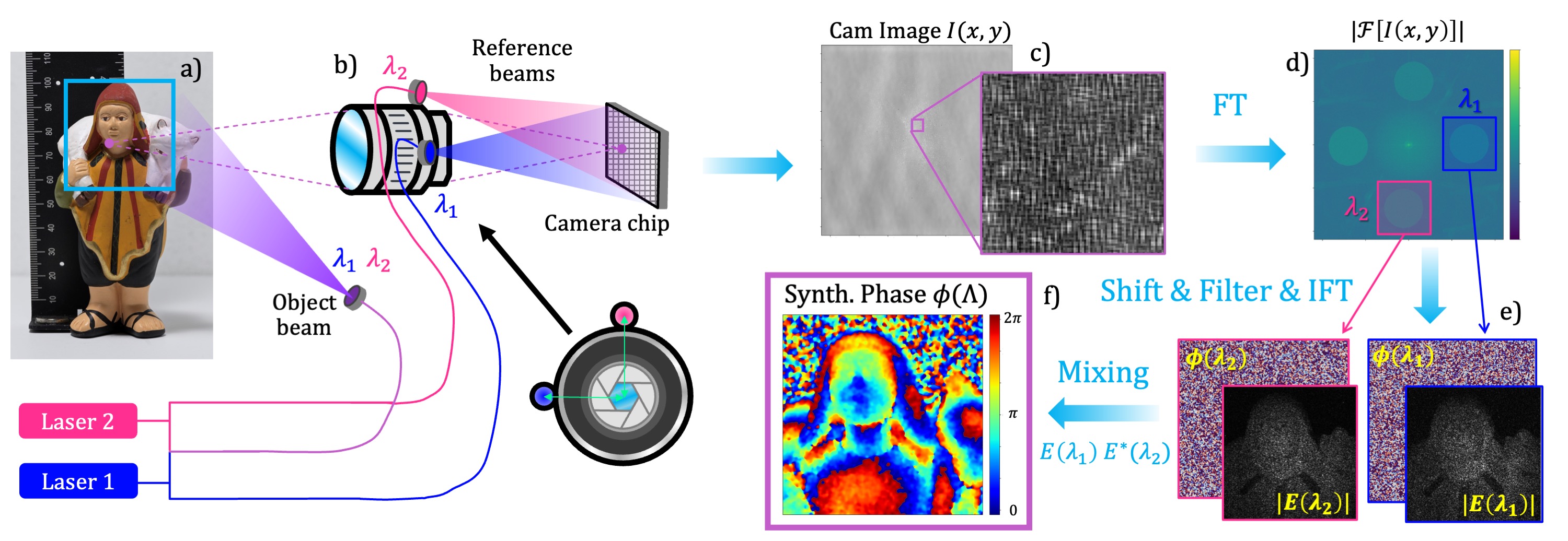}

   \caption{Schematic setup and computational procedure of our novel single-shot ToF camera system. (a) Object (the camera FoV is  $66 mm \times 66 mm$). (b) The lens system images the object onto a CCD/CMOS chip. An object beam at two wavelengths illuminates the object and two reference beams at two different wavelengths directly illuminate the chip at two different (vertical and horizontal) angles. (c) The captured image I(x,y) along with a zoom-in window. The image consists of speckles that are overlayed with crossed fringes. (d) Fourier transform of the captured image. (e) By shifting, filtering, and inverse Fourier transform, the speckled complex fields (amplitude and phase) at both optical wavelengths (see section \ref{sec:math}) can be recovered from only one camera image. (f) Using the SWI principle, we generate the synthetic phasemap $\phi(\Lambda)$ to calculate the 3D depth map of the object.}
   \label{fig:all}
\end{figure*}

\subsection{ToF camera concept}

Our ToF camera concept is based on Synthetic Wavelength Interferometry (SWI), a method for multifrequency interferometry  \cite{fercher1, fercher2, dandliker1, dandliker2, Groot1, Groot2}. SWI exploits the spectral diversity of multiple measurements at multiple wavelengths to image objects with optically rough surfaces. The basic principle is summarized as follows: An optically rough surface is illuminated with coherent light (at wavelength $\lambda_1$), and the complex field $E(\lambda_1)$ scattered off the object's surface is measured using an optical interferometer. This field exhibits strong wavefront aberrations due to the mentioned speckle. Optical pathlength information (i.e., the shape of the object) can not  be recovered from $E(\lambda_1)$ (see Fig. \ref{fig:all}e and Supp. Mat.). Eventually, the static object is again illuminated with a slightly different wavelength $\lambda_2$, and a new field $E(\lambda_2)$ is obtained. Assuming that both illumination sources originate from the exact same location (e.g., the same fiber tip), the fields $E(\lambda_1)$ and $E(\lambda_2)$ are subject to the same microscopic and macroscopic pathlength variations. After calculating the difference between their phasemaps $\phi(\lambda_1) - \phi(\lambda_2)$, the phase aberrations imparted by the microscopic pathlength variations cancel each other out. The phasemap difference only contains the macroscopic pathlength variations on the order of a ``Synthetic Wavelength" $\Lambda = \frac{\lambda_1 \lambda_2}{|\lambda_1-\lambda_2|}$ (see Fig. \ref{fig:all}f and Supp. Mat.).

If $\lambda_1$ and $\lambda_2$ are spaced close enough (see analysis in \cite{book, nature, arxiv_SWH}) the resulting ``synthetic field" $E(\Lambda)$ does not exhibit speckle artifacts and can be processed like a ``normal" ToF camera image or unspeckled optical interferogram. This means that the object's depth can be calculated by
\begin{equation}
\label{eq:depth}
z = \frac{1}{2} \frac{\phi(\Lambda)\cdot \Lambda}{2\pi} ~~.
\end{equation}

One possible way to calculate the synthetic field $E(\Lambda)$ is \textit{computational mixing} of $E(\lambda_1)$ and $E(\lambda_2)$,
\begin{equation*}
E(\Lambda) = E(\lambda_1)\cdot E^*(\lambda_2) = |E(\lambda_1)|\cdot |E(\lambda_2)| \cdot e^{i (\phi(\lambda_1)-\phi(\lambda_2))}
\end{equation*}
\begin{equation}
\label{eq:SW}
=|E(\lambda_1)| \cdot |E(\lambda_2)| \cdot e^{i \phi(\Lambda)}
\end{equation}

where $E^*(\lambda_2)$ denotes the complex conjugate of $E(\lambda_2)$. \\

Recent work \cite{PAMI, ICCP, nature, Ioannis} has introduced different ToF cameras for Computer Vision applications which are based on this synthetic wavelength principle. Despite the ability for ``full-field" measurements and measurements showing high-depth precision, the introduced systems had one significant drawback: at least two sequentially captured camera images were needed to reconstruct the object's depth maps. One would capture the field $E(\lambda_1)$ and then repeat the measurement using a different wavelength to obtain $E(\lambda_2)$. This approach works for static scenes but completely fails for dynamic scenes. If the object moves between the two sequential images, the optical pathlength between both images will differ. Therefore, the correlation between both speckle fields (needed to calculate the synthetic wavelength term) is lost. It turns out that this problem is much more severe than ``conventional" motion artifacts (such as those from structured light imaging systems), as minimal object movements can induce significant changes in the observed speckle patterns.  This presents a severe drawback because many real-world applications include scenes that naturally move at least on a microscopic scale (e.g., in medical imaging, autonomous driving, or AR/VR).

\subsection{Proposed method}

\begin{figure*}[t]
  \centering
   \includegraphics[width=\linewidth]{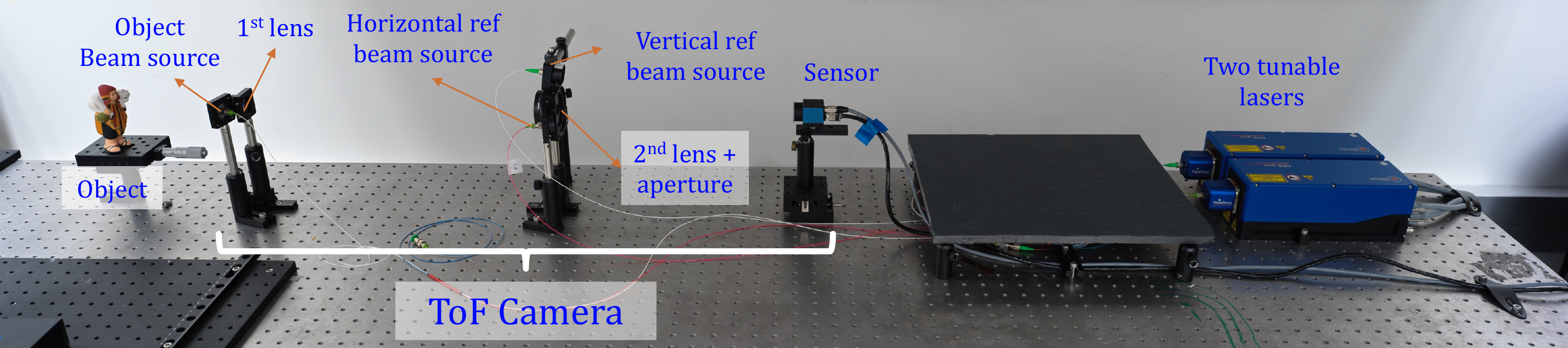}

   \caption{Picture of our lab setup. Two identical tunable lasers  emit light in the NIR range at around 850 nm. We use optical fibers and fiber beam splitters (BS) to direct the beams flexibly. We employ a lens system plus aperture to image the object into the sensor array.}
   \label{fig:setup}
\end{figure*}

The significant limitation of current SWI principles has stimulated the idea for this contribution. Our novel idea draws inspiration from spatial heterodyning, off-axis holography, and single-sideband demodulation procedures \cite{demodulation_essential, demodulation1, demodulation2, demodulation_highangle}. Our new setup configuration (see Fig. \ref{fig:all}a and \ref{fig:all}b) only uses standard CCD/CMOS camera technology and can operate in single-shot. Our approach can be summarized as follows: We couple a portion of two laser beams (at two different wavelengths $\lambda_1$ and $\lambda_2$) together, which forms the object beam. The other portions of both beams remain separated and form two reference beams (see Fig. \ref{fig:all}b). The reference beams are arranged so that they directly illuminate the camera chip at an angle. One reference beam (e.g., $\lambda_1$) encloses an angle with the horizontal x-axis, while the other reference beam (e.g., $\lambda_2$) encloses an angle with the vertical y-axis of the detector. 

When the object beam is scattered off the object surface, it forms a speckle field at each wavelength: $E(\lambda_1)$ and $E(\lambda_2)$. These two speckle fields are incident on the detector and interfere with the reference beams. The speckle field at $\lambda_1$ produces a static interference pattern with the  $\lambda_1$-reference beam (vertical fringes), while the speckle field at $\lambda_2$ produces a static interference pattern with the  $\lambda_2$-reference beam (horizontal fringes). In other words, the camera image $I(x,y)$ (Fig. \ref{fig:all}c) shows speckles that are overlayed by crossed fringes. For the sake of completeness, it should be mentioned that interference also happens between the fields at $\lambda_1$ and $\lambda_2$. These interferences result in temporally oscillating fringes which oscillate much faster than the integration time of our used camera and hence are not further discussed here (see Supplementary Material for details).

After acquiring the image $I(x,y)$ in single-shot, the complex speckle fields (phase and amplitude) $E(\lambda_1)$ and $E(\lambda_2)$ required to form the synthetic field  $E(\Lambda) = E(\lambda_1) \cdot E^*(\lambda_2)$ are retrieved via computational demodulation in the Fourier domain: Fig. \ref{fig:all}d shows the 2D Fourier transform of the captured image, $\mathcal{F}[I(x,y)]$, where five spectral regions with high energy can be distinguished. The central region represents the DC component. The horizontal regions (left and right) are centered around the spatial carrier frequencies of the vertical fringes, which appear due to the interferences between the  $\lambda_1$-reference beam and the object field $E(\lambda_1)$. Similarly, the vertical regions (up and down) are centered around the carrier frequencies for the horizontal fringes produced by the reference and object field at $\lambda_2$.  \\

Eventually, the complex field  $E(\lambda_1)$ is retrieved by performing the following operations:
\begin{itemize}

\setlength{\itemsep}{0pt}%

\item \underline{Find/evaluate the carrier frequency} $f_1$ for the respective spectrum in the Fourier domain (blue box in Fig.~\ref{fig:all}d).

\item \underline{Shift} the Fourier spectrum to set the evaluated carrier frequency $f_1$ as the new center frequency (see Supp. Mat. for further details).

\item \underline{Filter} the spectrum (either with a Hanning window or Gaussian kernel) so that only the frequency band around the new center frequency remains. The resulting filtered spectrum is denoted by $\mathcal{F}_\text{hor}[I(x,y)]$. 

\item An \underline{Inverse Fourier Transform} (IFT) of $\mathcal{F}_\text{hor}[I(x,y)]$ eventually delivers the phase $\phi(\lambda_1)$ and amplitude $|E(\lambda_1)|$ of the complex field $E(\lambda_1)$ (see Fig. \ref{fig:all}e and references \cite{demodulation_essential, demodulation1, demodulation2}):
\begin{equation}
\mathcal{F}^{-1} \{ \mathcal{F}_\text{hor} [I(x,y)] \} \propto |E(x,y;\lambda_1 )| \exp{(i \phi(x,y;\lambda_1))}
\end{equation}

\end{itemize}

The complex field $E(\lambda_2)$ is retrieved from the \textit{same image} in an analog fashion, using the vertically arranged regions in the Fourier spectrum. Eventually, the synthetic field $E(\Lambda)$ is formed via Eq. \ref{eq:SW}, the synthetic phase $\phi(\Lambda)$ is extracted, and the depth map of the object is calculated via Eq. \ref{eq:depth}. More details and mathematical calculations can be found in the Supp. Mat.

\section{Experiments and Data Processing}
\label{sec:Experiments}

\begin{figure*}[t]
  \centering
   \includegraphics[width=\linewidth]{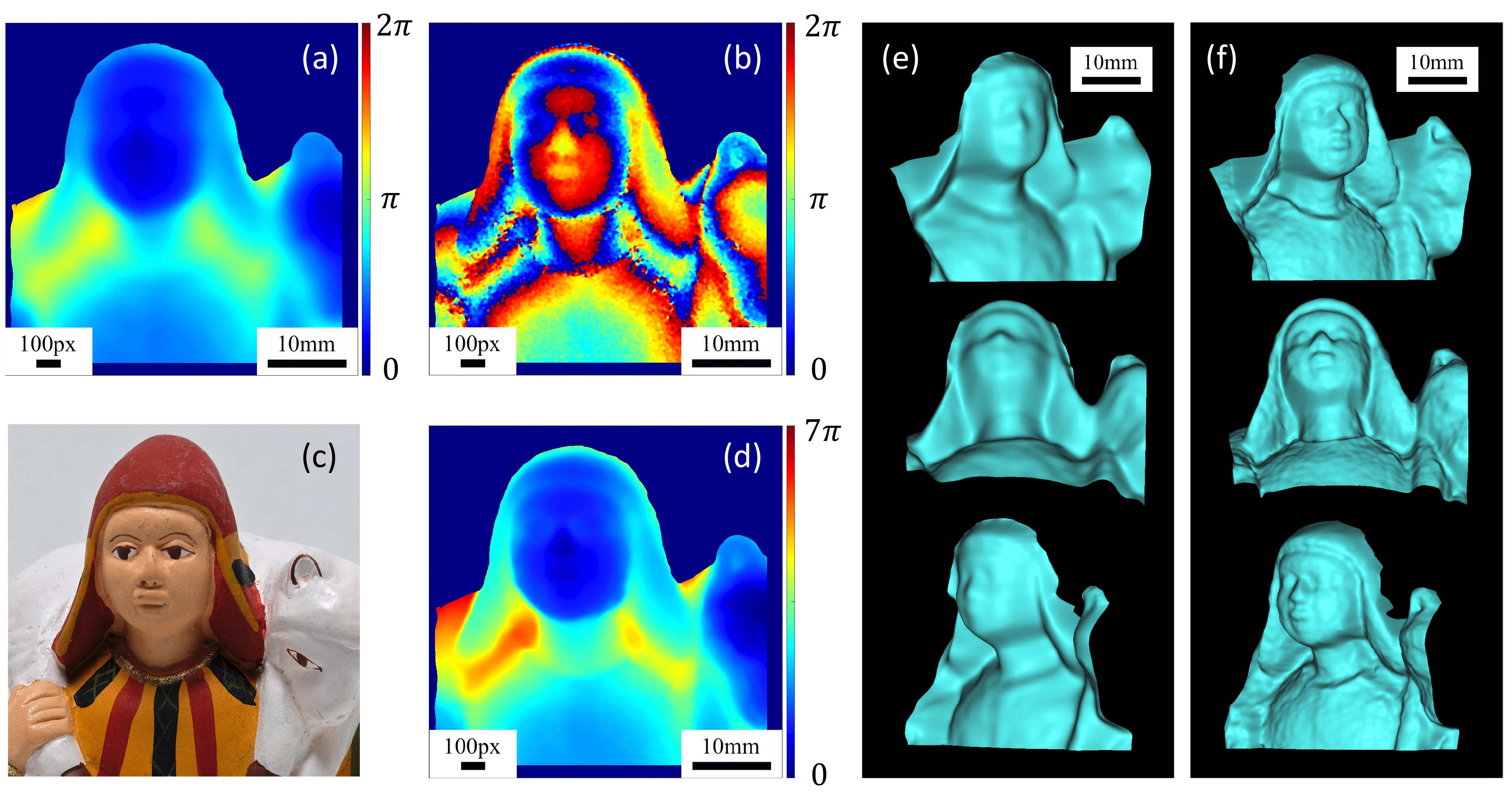}

   \caption{Double-shot acquisition mode. The measured object is a painted clay figure of approximately 10 cm height (see Fig \ref{fig:all}a), and the field of view is $66\times 66$ mm$^2$: (a) Acquired synthetic phase map  $\phi(\Lambda = 45\text{mm})$. (b) Wrapped phasemap $\phi(\Lambda = 10 \text{mm})$. (c) Image of the small object cropped to the actual FoV. (d) Phasemap $\phi(\Lambda = 10 \text{mm})$, unwrapped with our multi-frequency unwrapping algorithm. (e) and (f) 3D model of the object calculated from  $\phi(\Lambda = 45\text{mm})$ (e) and from unwrapped phasemap $\phi^\text{unwrap}(\Lambda =10\text{mm})$.}
   \label{fig:double}
\end{figure*}

\subsection{Experimental setup}

We first discuss the specific configuration of the setup used for our experiments, shown in Fig. \ref{fig:setup}. The optical system consists of a lens system that focuses the imaged object onto the camera chip. To incorporate the reference arms, it is required to leave a few centimeters between the sensor chip and its closest lens. This increases the back focal length of the system, affecting the Field of View (FoV) of the camera and making this first optical prototype system less compact. Further discussions on how to make the imaging system more compact are provided in section \ref{sec:discussions}.

It should also be mentioned that multiple possible setup configurations exist. As mentioned before, we use a reference source aligned on the horizontal axis and another on the vertical axis (see Fig. \ref{fig:setup}). However, other configurations of spatially separated reference beams are possible as well. For instance, we could place two reference beams along one single axis enclosing different angles. The only essential condition is that the Fourier transform of the captured image (see Fig. \ref{fig:all}d) must have separated spectral regions without overlap (aliasing) in distinct separated locations. 

To facilitate the reference and object beams at $\lambda_1$ and $\lambda_2$, two highly coherent light beams produced by two tunable lasers (\textit{Toptica DFB pro} devices) are flexibly directed by optical fibers. We separate the light emitted from each laser into the object and reference arms using a fiber splitter, where 90\% of the power goes to the object and 10\% to the reference. In addition, we combine the two object arms so that the resulting object source contains both wavelengths, $\lambda_1$ and $\lambda_2$ (see Supplementary Figure 2).

Although the object beam has much more intensity than the two references, we must consider that the light scattered off the object's surface that passes through the optical system is considerably less intense than the initial object beam. ND filters were used on the reference beams to match the intensity that reaches the sensor from the references and the object beam. When they match, the interference fringes have maximum contrast: destructive interferences with almost zero intensity and constructive fringes with four times the intensity of each beam \cite{hecht}. 

Additionally, we place an aperture right behind the objective lens (see Fig. \ref{fig:all}b) to control the size of the subjective speckles at the sensor location. This is essential, as our method requires several interference fringes within one speckle, so the speckle size needs to be controllable.  Respective trade-offs related to the aperture size and the lateral resolution of our camera system will be discussed in section \ref{sec:discussions}.

In the following, we use our described setup prototype to capture 3D images and 3D videos of different objects, using different single-shot and double-shot acquisition modes.

\subsection{Double-shot acquisition}

The ``double-shot" acquisition mode still captures the fields $E(\lambda_1)$ and $ E(\lambda_2)$ at both optical wavelengths sequentially. This means that \textit{only one} reference arm (Fig. \ref{fig:all}b) and \textit{only one} direction in the Fourier spectrum (Fig. \ref{fig:all}d) is exploited, and \textit{two} camera images are required to calculate the synthetic field $E(\Lambda)$ (hence the word ``double-shot"). However, it should be emphasized that each optical field $E(\lambda_1)$ or $ E(\lambda_2)$ itself is acquired in \textit{single-shot}, which is in stark contrast to conventional interferometry that normally relies on phase shifting and multiple exposures to capture the optical field.

Fig. \ref{fig:double} shows first double-shot acquisition results. The synthetic phasemap shown in Fig. \ref{fig:double}a was acquired at a synthetic wavelength $\Lambda = 45$ mm. The respective calculated 3D model is shown in Fig. \ref{fig:double}e from three different perspectives. Measurements at smaller synthetic wavelengths yield higher depth precision \cite{book, ICCP, PAMI}. However, for measurements at synthetic wavelengths $\Lambda$ smaller than 2x the object depth, the reconstructed synthetic phasemaps observe phase wrapping. To solve this problem, we applied a multi-frequency unwrapping algorithm to recover unwrapped phasemaps at a small synthetic wavelength using \textit{only a set of wrapped phasemaps} at different synthetic wavelengths. This means that a  "guidance phase map" at a large synthetic wavelength that does not observe wrapping is not required. Our applied approach works as follows: From two synthetic fields $E(\Lambda_1)$ and $E(\Lambda_2)$ with wrapped phasemaps, we calculate the beat-note field $E(\Lambda_B) = E(\Lambda_1) \cdot E^*(\Lambda_2)$, which could also be seen as the ``synthetic-synthetic field". If $E(\Lambda_1)$ and $E(\Lambda_2)$ are chosen correctly, the respective phasemap  $\phi(\Lambda_B)$ is not wrapped and can be now used as a guidance to unwrap fields at smaller synthetic wavelengths via the standard phase unwrapping procedure \cite{PAMI_unwrap1, PAMI_unwrap2}. Of course, this requires acquiring additional optical fields $E(\lambda_i)$ to form multiple synthetic fields at different synthetic wavelengths.

Fig. \ref{fig:double}b displays a wrapped synthetic phasemap of the object acquired at a synthetic wavelength of $\Lambda = 10$ mm. Fig.~\ref{fig:double}d shows the same phasemap, unwrapped with the procedure described above. The respective rendered 3D model is shown in Fig. \ref{fig:double}f. As expected, the 3D data display much higher depth resolution and details of the object surface. Fine details like the eyes or nose of the figure can be resolved with impressive quality (see scalebar for size comparison). Further results for the double-shot experiment are shown in the Supplementary Material (we encourage the reader to look at the measurement of the clay pot). It should also be mentioned that different smoothing operations are applied after each unwrapping step to facilitate ``smoother" guidance phasemaps. The exact smoothing parameters are specified in Supplementary Material.

\begin{figure*}[t]
  \centering
   \includegraphics[width=\linewidth]{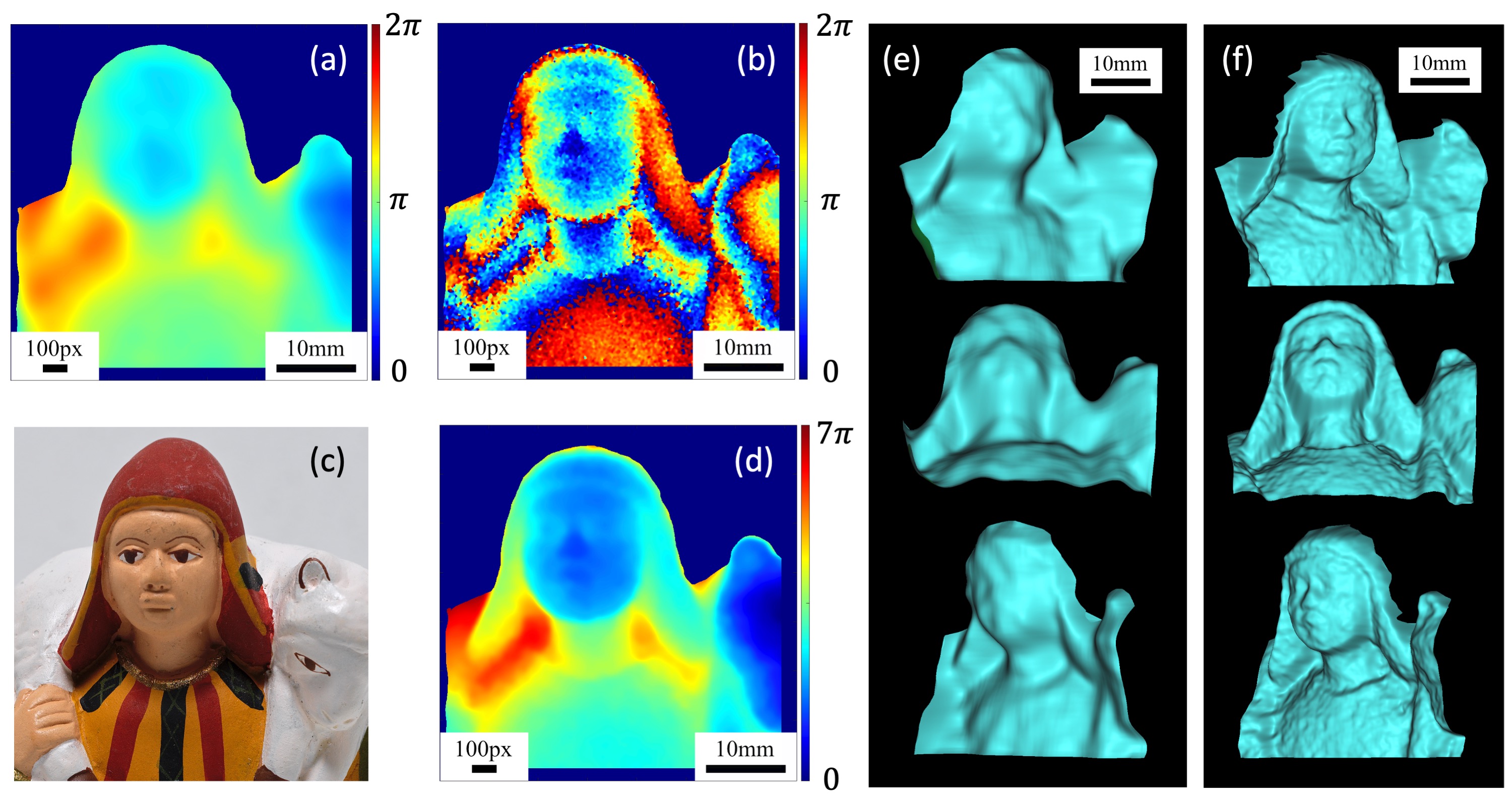}

   \caption{Single-shot acquisition mode. For the sake of comparability, the same object as in Fig. \ref{fig:double} is measured.  (a) Acquired synthetic phase map  $\phi(\Lambda = 50\text{mm})$. (b) Wrapped phasemap $\phi(\Lambda = 10 \text{mm})$. (c) Image of the small object cropped to the actual FoV. (d) Phasemap $\phi(\Lambda = 10 \text{mm})$, unwrapped with our multi-frequency unwrapping algorithm. (e) and (f) 3D model of the object calculated from  $\phi(\Lambda = 50\text{mm})$ (e) and from unwrapped phasemap $\phi^\text{unwrap}(\Lambda =10\text{mm})$.}
   \label{fig:single}
\end{figure*}

\subsection{Single-shot acquisition}

The ``single-shot" acquisition mode captures the synthetic field $E(\Lambda)$ with \textit{only one} camera image, using the full Fourier demodulation procedure (with two reference beams) described in section \ref{sec:math}. The results in Fig. \ref{fig:single} are intentionally presented in an analog fashion to Fig. \ref{fig:double} to allow for comparing both acquisition modes.

Figure \ref{fig:single}a shows a single-shot phasemap captured at a synthetic wavelength of $\Lambda= 50$ mm - large enough that the phasemap is ``unique", i.e., not subject to phase wrapping. The respective 3D model is shown in Fig. \ref{fig:single}e from three different perspectives. Figure \ref{fig:single}b shows a wrapped single-shot phasemap captured at $\Lambda= 10$ mm. The same phasemap processed with the unwrapping procedure described above is displayed in Fig. \ref{fig:single}d, and the respective 3D model is shown in Fig. \ref{fig:single}f. It can be seen that the quality and detail richness of the measurement does not significantly decrease for the ``single-shot" compared to the previously shown ``double-shot" method. \\

For a quantitative depth precision evaluation of our camera and a comparison between single-shot and double-shot acquisition, we evaluated single-shot and double-shot measurements of a planar surface (cardboard) at different synthetic wavelengths. A small region of the cardboard (roughly $23mm \times 23mm$) has been picked for evaluation. After subtracting a best-fit plane, the standard deviation of the (unfiltered) point cloud was calculated. This value represents the depth precision $\delta z$ of the respective measurement. The results are displayed in Tab. \ref{tab:table1}: It can be seen that the single-shot measurements achieve roughly the same depth precision as their double-shot counterparts at the same synthetic wavelength. For $\Lambda \leq 3$ mm, our camera achieves sub-mm precision with a best-reported precision of only \textbf{ $\delta z_{single} = 330 \mu m$ } for $\Lambda = 1$ mm. This value outperforms the precision of conventional ToF cameras roughly by a factor of \textbf{100x}. \\

\begin{table}
    \centering
    \begin{tabular}{|c|c|c|c|c|c|}
        \hline   $\Lambda$ [mm] & 40 & 10 & 5 & 3 & 1 \\
        \hline   $\delta z_\text{double}$ [mm] & 5.30 & 1.95 & 1.14 & 0.82 & 0.43 \\
        \hline   $\delta z_\text{single}$ [mm] & 5.56 & 1.78 & 1.64 & 0.79 & 0.33 \\
        \hline
    \end{tabular}
    \caption{Depth precision analysis comparing the double-shot with the single-shot method.}
  \label{tab:table1}
\end{table}

\begin{figure}[b]
  \centering
   \includegraphics[width=\linewidth]{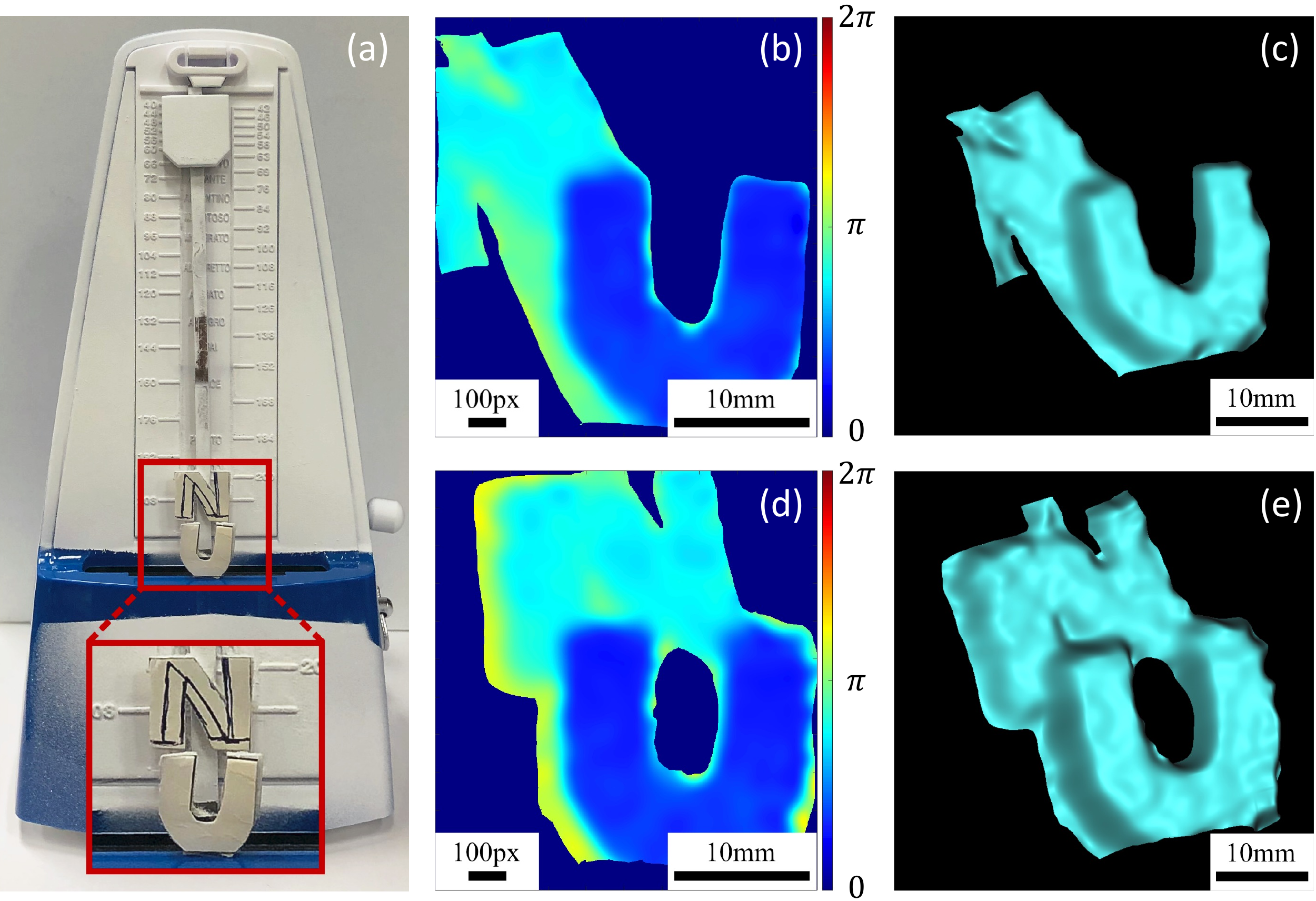}
   \caption{Single-shot measurements of a dynamic object. (a) Object: Metronome sprayed with a scattering powder. (b, d) Phasemaps $\phi(\Lambda = 30 \text{mm})$ obtained from two different video frames. (c, e) Corresponding 3D models.}
   \label{fig:metronome}
\end{figure}

\begin{figure*}[t]
  \centering
   \includegraphics[width=\linewidth]{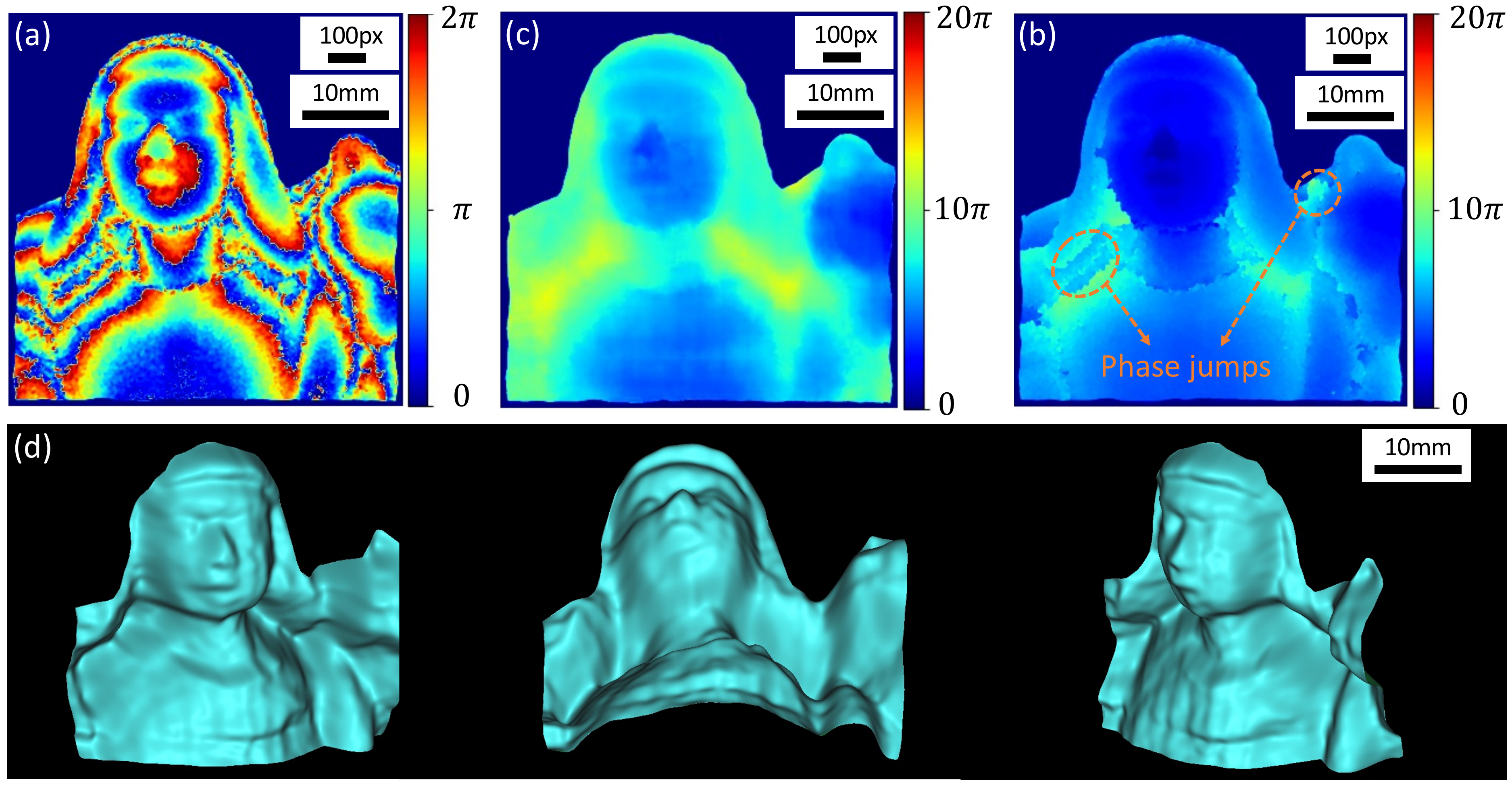}
   \caption{Single-shot phase unwrapping using Deep Learning (DL). (a) Acquired wrapped synthetic phase map $\phi(\Lambda = 5 \text{mm})$. (b) Unwrapped phase map predicted by our DL algorithm, $\phi^\text{unwrap}(\Lambda = 5 \text{mm})$. (c) Comparison: Phasemap obtained from another single-shot unwrapping procedure \cite{conventional_unwrap1, conventional_unwrap2} (not DL-based). Unwrapping failed as phase jumps are still present.
   (d) 3D model calculated from our DL-based unwrapped phasemap (b).}
   \label{fig:DL}
\end{figure*}

To further demonstrate the single-shot capability of our novel Synthetic Wavelength Interferometry (SWI) ToF camera, we recorded a 3D video of a moving object. Each video frame consists of a single-shot measurement of the object at a synthetic wavelength of $\Lambda = 30$ mm. The object consists of a metronome whose pendulum moves left-to-right and then back right-to-left with multiple oscillations. To make the movement more obvious, two letters are attached to the metronome - one letter at the pendulum and one letter at the metronome body. During oscillation, the back letter moves laterally with respect to the front letter. 

Figure  \ref{fig:metronome} b and d show two synthetic phasemaps from two different time instances of the video. The respective 3D reconstructions are shown in Fig. \ref{fig:metronome} c and e. The full video of the captured phase map and 3D reconstruction can be seen in link \cite{video}. We emphasize again that such a 3D video sequence would not have been possible with earlier implementations of Synthetic Wavelength Interferometry for Computer Vision applications shown, e.g., in \cite{PAMI, ICCP, nature, Ioannis}.

In light of the encouraging single-shot results that have been demonstrated so far, one could plausibly argue that the ``unwrapped single-shot'' results shown in Fig. \ref{fig:single} are \textit{not} genuinely single-shot, as we needed multiple sequentially captured single-shot phasemaps to apply the multifrequency unwrapping algorithm. And indeed, if we want to exploit the main feature of our novel camera (namely precise single-shot 3D acquisition of dynamic objects), multifrequency phase unwrapping from sequential images would not be possible\footnote{It should be emphasized, however, that the discussed motion artifact problem is now reduced to ``conventional motion artifacts'' as, e.g., observed in multi-shot structured light principles. This is because the multifrequency unwrapping with two single-shot synthetic phase maps does not require the speckle patterns to be correlated between the different time instances (see the previous discussion).}. For this reason, we have started to develop a Deep Learning (DL) based algorithm to unwrap our captured (wrapped) single-shot phase maps at smaller synthetic wavelengths. 

Compared to other SOTA DL-based phase unwrapping procedures \cite{unwrap_heide1, unwrap_heide2, unwrap_probabilistic, unwrap_maxlikelihood}, our algorithm is specifically tailored to our acquired data structures and integrates seamlessly into our data processing pipeline. We defined a Convolutional Neural Network (CNN) architecture combined with an LSTM module \cite{unwrap_DL, unwrap_intro, unwrap_CNN, unwrap_LSTM}. The network is trained using measured and simulated wrapped phasemaps at different synthetic wavelengths as input and unwrapped phasemaps as output. The experimentally acquired unwrapped phase maps for the training process were obtained using the multifrequency unwrapping  algorithm described above. The supervised learning machine algorithm has a loss function that measures the mean squared error (MSE) between the real output image and the output that the CNN architecture generates from the input data. During the training process, the parameters of the CNN are optimized so that the model produces accurate unwrapped output phasemaps out of the wrapped input phase maps. Once the training is complete, we apply the model to a newly measured  wrapped phasemap to predict the unwrapped version. Neither the measured phasemap nor any other measurement of the evaluated object surface have been seen before by the algorithm.

Although not the main focus of this paper and currently a work in progress, we here report the first single-shot measurements unwrapped with our DL-based algorithm in Fig. \ref{fig:DL}. It can be seen that the algorithm correctly unwraps and reconstructs the overall shape of the object (see Fig. \ref{fig:DL}b). However, fine details like eyes and nose are reconstructed not as well as with the multifrequency approach (see Figs. \ref{fig:double}  and \ref{fig:single}). For comparison, we also attempted to unwrap the phasemap with an often-used single-shot unwrapping procedure (not DL-based) described in \cite{conventional_unwrap1, conventional_unwrap2}. The result is shown in Fig. \ref{fig:DL}c. It can be seen that this other method was not able to unwrap our synthetic phasemap correctly, and phase jumps are still present in the result.

\subsection{Single-shot NLoS imaging through scattering scenes}

\begin{figure}[b]
  \centering
   \includegraphics[width=\linewidth]{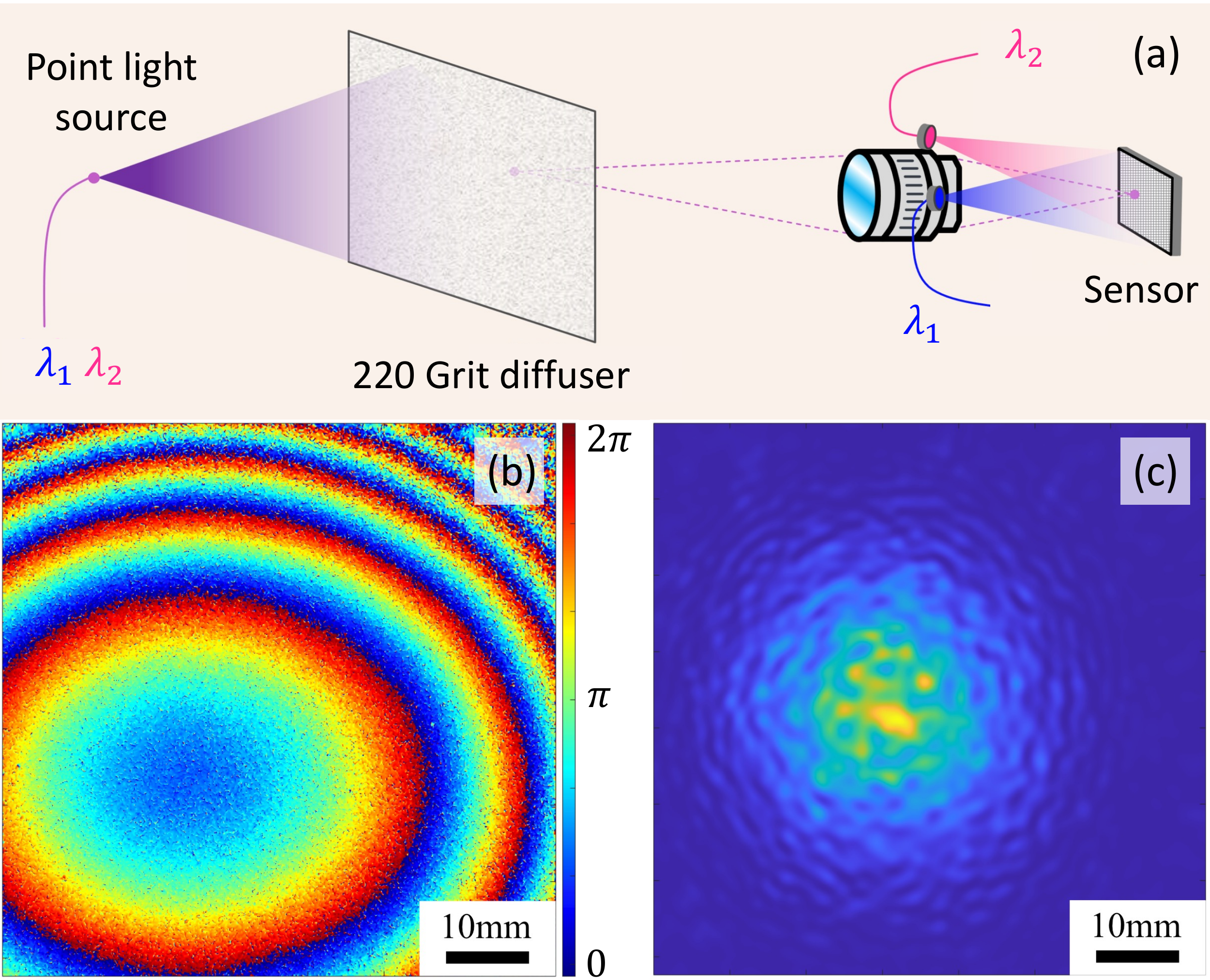}

   \caption{Full-field, single-shot Non-Light-of-Sight (NLoS) imaging. (a) Schematic of the experiment. (b) Synthetic phasemap $\phi(\Lambda = 1 \text{mm})$ of the object's hologram at the diffuser location. (c) The amplitude of backpropagated object (the point light  source).}
   \label{fig:NLoS}
\end{figure}

So far, this paper was focused on "Line-of-Sight" (LoS) measurements, where the object is directly visible and can be directly imaged onto the sensor. However, the same technique can be used to image objects hidden from direct view, e.g., behind a scattering medium or around a corner. These so-called ``Non-Line-of-Sight'' (NLos) measurements are potentially  useful for many future applications in autonomous driving, medical diagnosis, or quality inspection of manufacturing products. For instance, we could think of an imaging system that could see through fog, that could see through human tissue to observe the inner organs, or that could look ``around corners'' to see some industrial machinery in operations.

Multi-shot NLoS measurements using Synthetic wavelength imaging principles have been demonstrated in \cite{nature}.  However, it turns out that the single-shot ability of our camera is particularly crucial for NLoS measurements - even if the hidden object does not move! The reason is that most scattering media, such as fog, smoke, turbid water, or even human tissue, are in constant microscopic motion. For the double-shot method introduced in \cite{nature}, this would lead to the already discussed decorrelation of speckle patterns which would result in a complete loss of information. In Fig. \ref{fig:NLoS}, we demonstrate a first single-shot NLoS measurement through a scattering scene. As shown in Fig. \ref{fig:NLoS}a, the object behind the scatterer is a simple point light source (fiber tip), which emits light at both used optical wavelengths $\lambda_1$ and $\lambda_2$. The scatterer is a 220-grit ground glass diffuser. The synthetic phasemap $\phi(\Lambda)$ (see Fig \ref{fig:NLoS}b) is captured at the diffuser surface, and the related field is computationally backpropagated at the synthetic wavelength to reconstruct the point source in the hidden volume behind the diffuser. We refer to \cite{nature} for details about the used reconstruction method. An image of the reconstructed point light source is shown in Fig. \ref{fig:NLoS}c. The diameter of the bright spot in the middle roughly matches the theoretical expectations about the reconstruction resolution of the point light source. Further explanations are provided in the supplementary material.

Although very basic, this experiment documents (to the best of our knowledge) the first ever SWI-based single-shot full-field measurement through scattering media.

\section{Summary, discussion,  and outlook}
\label{sec:discussions}

We first summarize the advantages of our novel method with respect to other ToF-based imaging systems. While standard (single-wavelength) interferometric ToF cameras are limited to specular objects, our novel system can image objects with an optically rough surface. Other existing multifrequency interferometric ToF cameras for Computer Vision applications were limited to static objects and/or used sophisticated sensors (e.g., PMDs \cite{cwam1}, SPADs \cite{SPAD}, or lock-in detectors \cite{lock-in}). In contrast, this work demonstrates a new setup design that enables high-precision measurements with conventional CCD/CMOS sensors. Our main contribution to the field is the introduction of our single-shot SWI method, which allows us to record videos of moving objects and reconstruct the dynamic 3D model at high precision. 

In addition, we have demonstrated ``unwrapped dual-shot measurements" and ``unwrapped single-shot measurements" with a depth precision of up to $330 \mu m$ - roughly 100x better than the precision of conventional ToF cameras. Due to this immense performance difference, we focused our experiments around demonstrating the capabilities of \textit{our} system and refrained from comparisons with conventional ToF camera systems (such as \textit{Microsoft Kinect V2} \cite{kinect}). We hope that it became obvious to the reader that such systems would not have been able to sufficiently resolve any of our small measured objects at all. 

We also want to mention that our single-shot method is not only limited to the simultaneous measurement of two optical wavelengths, i.e., one synthetic wavelength. Indeed, we could theoretically operate three  lasers simultaneously and illuminate the object with $\lambda_1$, $\lambda_2$, and $\lambda_3$. For instance, we could  place each of the three reference beams with a 60-degree difference around the aperture. With this procedure, one could retrieve the object beams $E(\lambda_1)$, $E(\lambda_2)$, and $E(\lambda_3)$ from a single-shot image following our proposed method (see section 2.2). This new configuration enables the construction of three different synthetic fields: $E(\Lambda_{12})$, $E(\Lambda_{23})$, and $E(\Lambda_{13})$. Consequently, one can then use the "standard" multi-frequency unwrapping technique to retrieve a precise depth map from an unwrapped phase map at a small synthetic wavelength \textit{purely in single-shot}.

Of course, the principle can be extended to even more optical wavelengths as long as the different regions in the Fourier domain stay separated. Further developments in this direction will be part of our future work.\\

Despite the promising first results shown in this paper, our imaging system is not without limitations. On the one hand, there is a fundamental limitation related to the aperture size of the optical system. A wider aperture size would allow higher spatial frequencies from the object beam to pass, resulting in higher lateral resolution during the imaging process but also increasing the size of the spectral regions in the Fourier domain. According to this tradeoff, we must carefully calibrate the aperture size so that the spectral regions in the Fourier domain do not overlap. Besides the tradeoff mentioned above, a small aperture also leads to decreased light throughput and hence a longer required exposure time. This issue becomes more severe when we record videos. We need a short exposure time to capture unblurred images of moving objects, which decreases the amount of light arriving at the sensor. The ``optimal" aperture size is reached shortly before the spectral regions overlap in the Fourier domain.

Another limitation of our current system is its limited field of view (FoV). The back focal length of a camera is defined by the distance between the camera sensor and the closest lens from the optical system (see Fig. \ref{fig:setup}). As we need to place the reference beams pointing directly toward the camera sensor chip, we have to increase the back focal length of our camera. A short back focal length would imply a larger angle between the reference beams and the optical axis, which reduces the spacing between the fringes (see Fig. \ref{fig:all}c). As we are already operating near the sampling limit (with about three pixels per fringe), the back focal length of our system cannot be shortened, and the total size of the setup becomes large. In the future, we will explore different solution approaches to this problem which can be found in optical metrology literature (e.g., in \cite{demodulation_highangle}).

\section{Conclusions}
\label{sec:conclusions}

We have developed a novel ToF single-shot imaging system based on Synthetic Wavelength Interferometry (SWI). Our contributions extend previous works on SWI-based ToF cameras, which were either limited to static objects or used sophisticated sensor architectures. We have shown the design of the new camera prototype and presented the computational procedure to obtain high-precision depth maps for several different introduced acquisition modes. We have tested our imaging system by measuring different static and dynamic objects and have acquired a ``single-shot 3D video" of a moving object. Ultimately, we have exposed another potential new application of our technology by using our camera for a first-ever SWI-based single-shot NLoS measurement.

In the future, we hope that our introduced technique becomes part of a new wave of imaging devices that allow for completely new procedures and methods on how we perceive and interact with our environment.

\begin{acknowledgements}
\noindent
The authors thank Prasanna V. Rangarajan and Muralidhar M. Balaji for their thoughtful comments and for the vivid discussions. The authors thank Wildaline Serin for her diligent proofreading of the manuscript.

\end{acknowledgements}

\section{Bibliography}
\bibliography{zHenriquesLab-Mendeley}


\end{document}